\documentclass{isprs}
\usepackage{setspace}
\usepackage{geometry} 
\usepackage{epstopdf}
\usepackage[labelsep=period]{caption}  
\usepackage[british]{babel} 
\usepackage[hang]{footmisc}
\usepackage{multirow}

\usepackage{amsmath}   
\usepackage{amssymb}   
\usepackage{bm}   
\usepackage{booktabs}
\usepackage{float}
\usepackage{dblfloatfix}

\usepackage{graphicx}
\usepackage{subcaption}

\usepackage{listings}
\usepackage{xcolor}
\usepackage{comment}
\usepackage{tikz}
\usetikzlibrary{arrows.meta,positioning,fit,backgrounds,matrix,shapes.geometric,shapes.symbols,angles,quotes,calc}

\usepackage{url}

\usepackage{minted}
\setminted{
    fontsize=\footnotesize,
    baselinestretch=1.1,
    frame=none,
    framesep=2mm,
    bgcolor=gray!5,
    breaklines=true,        
    breakanywhere=false
}

\usepackage[authoryear]{natbib}

\begin{document}

\title{CityLLM: A framework for natural-language querying of semantic 3D city models}
\date{}


\author{Rabindra Lamsal\textsuperscript{1}, Sisi Zlatanova\textsuperscript{1}, Johnson Xuesong Shen\textsuperscript{2}}

\address{
\textsuperscript{1}GRID Lab, School of Built Environment, UNSW Sydney, NSW 2052, Australia - (r.lamsal,s.zlatanova)@unsw.edu.au\\
\textsuperscript{2}School of Civil and Environmental Engineering, UNSW Sydney, NSW 2052, Australia - x.shen@unsw.edu.au
}



\abstract{
Semantic 3D city models provide rich geometric and semantic information, but remain challenging for non-experts and interdisciplinary researchers to access and query due to their complex structures and specialized data formats. To address this issue, we present CityLLM, a framework for natural-language querying of semantic 3D city models alongside complementary urban datasets. The framework combines spatial and graph databases within an LLM-based workflow that supports iterative query refinement and cross-database chaining. We evaluate CityLLM on a CityJSON dataset of Rotterdam (853 LoD2 buildings) using GPT-OSS, Gemini 3.1, and GPT-5.4, along with selected variants, across multiple metrics: \textit{answer correctness}, \textit{visualization correctness}, \textit{query success}, and \textit{retry attempts}. A total of 54 natural-language queries are curated across four scenarios: spatial, graph, cross-database, and conversational. Results show strong overall performance, with answer correctness ranging from 85.2\% to 100\%, visualization correctness from 92.9\% to 100\%, a 100\% query success rate, and fewer than three retries across all 54 queries. Overall, the findings suggest that CityLLM provides a lightweight and extensible approach for conversational access to semantic 3D city data.
}


\maketitle


\section{Introduction}
Semantic 3D city models are a core component of urban digital twins and smart-city applications. CityGML \citep{groger2012citygml}, an OGC standard, provides a rich XML-based exchange format for representing 3D city models, capturing geometry, semantics, topology, and multiple Levels of Detail (LoDs) for urban objects such as buildings, transportation networks, water bodies, and land use \citep{kolbe2009representing}. CityJSON \citep{ledoux2019cityjson}, an alternative encoding of the same data model, offers a more compact, developer-friendly JSON representation suited to web-based and lightweight workflows. Together, these standards have enabled the creation and dissemination of large-scale semantic 3D city datasets\footnote{https://3d.bk.tudelft.nl/opendata/opencities/} that are increasingly used in both research and practical urban applications. 

In practice, however, these models remain difficult to access and query. Their hierarchical structure and spatial characteristics demand specialized knowledge of data schemas, spatial query languages, or dedicated GIS software; a barrier that grows when a city model must be queried alongside other urban data sources, such as street networks or amenities, that follow entirely different formats. Effective exploration therefore remains challenging for non-expert and interdisciplinary users alike, especially when a task requires drawing on multiple sources rather than the city model alone. As urban digital twins grow in scale and heterogeneity, more intuitive and accessible ways of interacting with this data are needed.

Recent advances in large language models (LLMs) have opened new possibilities for natural-language interaction with complex data \citep{achiam2023gpt}. LLMs have shown strong potential for translating natural-language questions into structured queries, coordinating external tools, and supporting multi-step reasoning. These capabilities are particularly relevant for geospatial applications, where users often need to express complex spatial questions or combine information from multiple data sources. For semantic 3D city models, LLM-based interfaces can lower this barrier by letting users ask questions in plain language instead of writing format-native queries (e.g., XQuery, SQL, Cypher). However, this is not straightforward to realize: these datasets combine geometric, semantic, and spatial-relationship information, demanding schema awareness and reliable query generation, and failures can arise when schema context is incomplete, when reasoning spans multiple steps, or when a query must integrate heterogeneous sources. Existing approaches have shown strong potential in this direction \citep{liu2025kcitychatbot,kanna2025automatic,kanna2025advanced}, but less attention has been given to lightweight, extensible workflows that coordinate complementary data representations of semantic 3D city models within a single conversational pipeline.

To address this, we present \textit{CityLLM}, a framework for natural-language querying of semantic 3D city models. The framework combines an LLM-based agent with two complementary data backends: CityJSON-derived data are stored in a lightweight spatial database for semantic and spatial queries, while street networks and amenities from OpenStreetMap are represented in a graph database for graph-based analysis. The agent interprets user queries, selects the appropriate backend(s), generates and executes queries, and returns responses that can be visualized in an interactive map. The workflow also supports iterative query refinement and multi-step query chaining across heterogeneous data sources.

The remainder of this paper is organized as follows. Section~\ref{sec:relatedwork} reviews the relevant literature, Section~\ref{sec:architecture} describes the proposed framework, Section~\ref{sec:evaluation} discusses the evaluation setup, Section~\ref{sec:results} presents the results and discussion, and Section~\ref{sec:conclusion} concludes the work.

\section{Literature Review}
\label{sec:relatedwork}
Recent advances in LLMs \citep{achiam2023gpt} are pushing the boundaries of geospatial research. Although the integration of LLMs with geospatial data is still in its early stages \citep{kanna2025advanced}, researchers are increasingly embedding LLMs into diverse AI workflows, ranging from natural-language interaction \citep{pan2026llm} to disaster geolocalization \citep{yin2025llm} and automated geospatial modeling \citep{liang2025geographrag}, among other emerging applications. In fact, recent works have begun to envision urban foundation models that place LLM-based agents at the core of city analytics. For example, \citep{zhang2024towards} outlined a taxonomy for urban data and discussed how foundation models can act as central agents for multimodal urban tasks, and \citep{xu2023urban} proposed an Urban Generative Intelligence platform that trained a specialized model ``CityGPT" with agents to simulate and reason about complex city environments. Such developments highlight the momentum toward integrating LLMs as core components of urban digital twins and decision-support systems. Concurrently, other research highlights the role of generative AI in smart cities and the metaverse; for instance, \citep{lifelo2024artificial} emphasized that advanced NLP capabilities of LLMs enable interactive conversational experiences within urban digital twins, enhancing user engagement in smart city applications.

One of the most immediate applications of LLMs in the geospatial domain is question answering (QA). For instance, \citep{feng2023geoqamap} introduced GeoQAMap (Geographic Question Answering), an LLM-based system that translates natural-language queries into SPARQL queries over the Wikidata knowledge base \citep{Wikidata} and visualizes the results on interactive maps. Similarly, \citep{deng2025question} developed Zaha, a QA system integrated with \textit{The World Avatar} knowledge graph \citep{akroyd2021universal}, which enables intuitive natural-language queries for smart city and urban planning data by unifying information from diverse sources. In a related context, \citep{santhanavanich2025ogc} presented OGC-AI, an LLM-based interface that allows users to interact with OGC web services in natural-language. These approaches show how conversational AI can lower the barrier for non-experts to access and analyze complex geospatial data through natural-language.

Particularly concerning semantic 3D city models, their rich but complex schema poses challenges for direct interaction by users. One of the notable works in making CityGML more interpretable was by \citep{hansson2024linked}, where CityGML data was represented as a virtual knowledge graph using its ontology and R2RML \citep{world2012r2rml} mappings. Furthermore, in the context of mapping, \citep{kanna2025automatic} proposed a system to automatically augment CityGML data with missing real-world attributes. Their approach involved extracting information from external sources (e.g., OpenStreetMap) and mapping it to corresponding CityGML features. First, the CityGML data is imported into a spatial relational database using 3DCityDB \citep{3DCityDatabase}, and an LLM agent generates and executes SQL insert statements to populate the newly extracted attributes into the CityGML schema.

Limited studies to date have focused on creating full conversational interfaces for semantic 3D city models. One example is KCityChatBot \citep{liu2025kcitychatbot}, which allows natural-language dialogue with a CityGML-based knowledge graph. Their method automatically transforms CityGML data into a property graph, representing each city object as a node with edges capturing the nested hierarchies and then uses a multi-agent LLM framework to handle user queries. In this pipeline, one agent interprets a user's question and generates a corresponding Cypher query, which is executed on a Neo4j graph database, and another agent formulates a natural-language answer based on the query results. Likewise, \citep{kanna2025automatic} implemented an LLM-based query agent that converts natural-language questions into SQL against a 3DCityDB database. Extending this approach, \citep{kanna2025advanced} presented an advanced LLM-driven interface for an urban digital twin that combines the semantic city model with time-series sensor data. In their framework, city objects and their dynamic observations are stored in the database, and the LLM acts as an orchestrator that can iteratively refine its queries. For example, given a user prompt, the LLM generates an SQL query (or calls an external tool) to fetch information from either the static CityGML data or the linked sensor data, and then progressively improves the response through follow-up queries or tool use, if needed. The results are presented in natural-language and can be visualized. This interactive system showcases the potential of combining LLMs with 3D city models and real-time data for conversational query and analysis.

Overall, these studies represent important progress toward conversational access to semantic 3D city and urban digital twin data, while also indicating complementary design directions. KCityChatBot \citep{liu2025kcitychatbot} demonstrates the value of a knowledge-graph-based interface for large-scale CityGML interaction, whereas \citep{kanna2025automatic} highlights LLM-supported enrichment and SQL-based access to CityGML workflows. The advanced interface in \citep{kanna2025advanced} further shows the potential of LLM orchestration across static city models and dynamic sensor streams. Building on these efforts, this work focuses on coordinating complementary data representations of semantic 3D city models within a single lightweight architecture, one that decides which data backend a user query needs, chains results across them when a query requires both, supports iterative query repair and exposes the generated queries and resulting visualizations through one conversational interface.

\begin{figure*}
    \centering
    \includegraphics[width=0.68\linewidth]{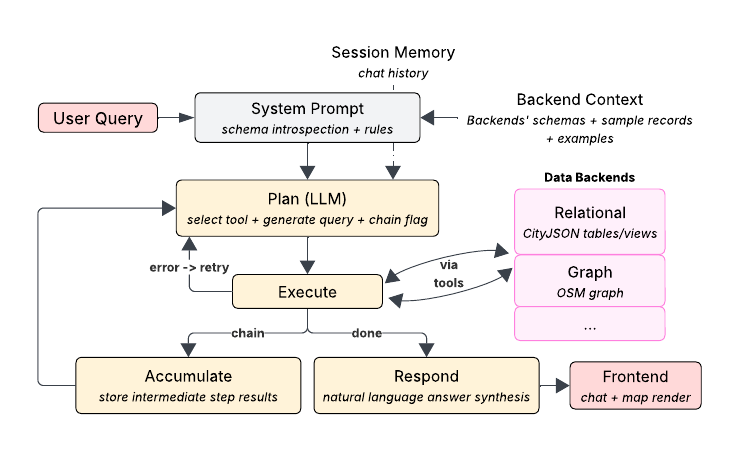}
    \caption{Architecture of the CityLLM framework.}
    \label{fig:workflow}
\end{figure*}

\section{Proposed Framework}
\label{sec:architecture}

The proposed framework, CityLLM, enables natural-language interaction with semantic 3D city models by integrating an LLM with spatial and graph databases. Its architecture is illustrated in Figure~\ref{fig:workflow}. The framework follows a modular design composed of three main components: (i) data backends, (ii) an LLM-based agent, and (iii) an interactive map-based interface. In this section, we describe each component, including its implementation and the underlying sub-components that together form the CityLLM framework.

\subsection{Data Backends}
\label{sec:data-preparation}

\begin{figure}
    \centering
    \includegraphics[width=0.85\linewidth]{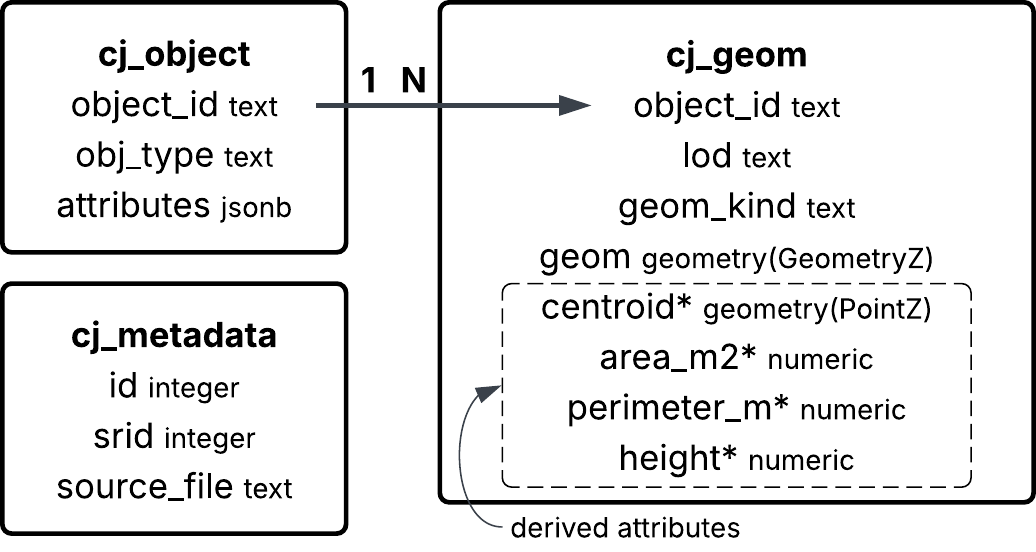}
    \caption{Relational database (PostgreSQL/PostGIS) schema.}
    \label{fig:postgres-schema}
\end{figure}

Semantic 3D city models in this study are represented in the CityJSON format. To support spatial querying, a CityJSON dataset is imported into a PostgreSQL/PostGIS database using a compact relational schema (as shown in Figure \ref{fig:postgres-schema}). The schema is intentionally simple, reducing structural complexity and avoiding excessive joins, which facilitates more reliable query generation by the LLM while minimizing output tokens and inference cost.

During preprocessing, the dataset is parsed and mapped to the relational schema. Geometries are stored using native spatial data types to support 3D operations and indexing, while semantic attributes are kept separately. Since a city object can have multiple geometries across different LoDs, object attributes are stored in one table and the corresponding geometries in another, linked via the shared \textit{object\_id}. The geometries are inserted using the dataset’s reference coordinate system. In addition, derived properties such as centroid, area, perimeter, and height are computed during preprocessing and stored alongside the geometries. This avoids repeated computation at query time and allows the LLM to directly reference these attributes when answering user queries. With this setup, spatial queries are executed using standard PostGIS functions (e.g., \textit{ST\_Intersects}, \textit{ST\_Distance}, \textit{ST\_DWithin}), supporting tasks such as proximity search, spatial filtering, and geometric analysis. In addition to the base tables, a view is created that combines object attributes with a single geometry per object, selected as the highest available LoD, for use in most LLM queries.

\begin{figure}
    \centering
    \includegraphics[width=1\linewidth]{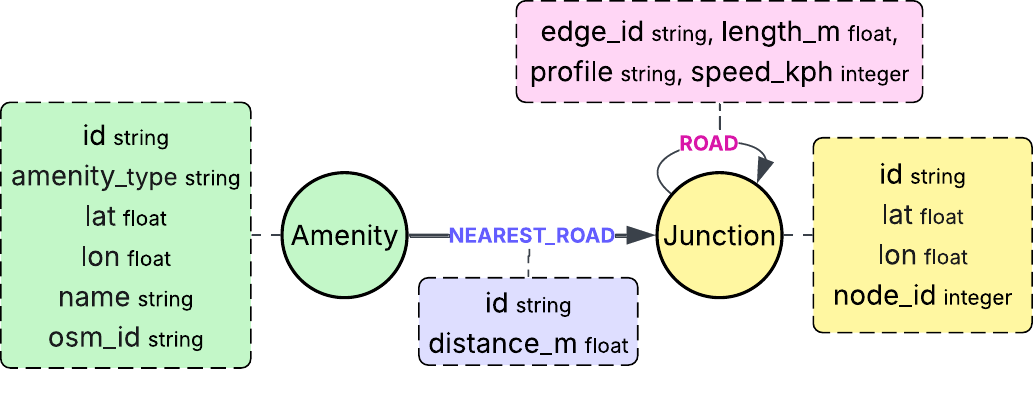}
    \caption{Graph database (Neo4j) schema.}
    \label{fig:graph-schema}
\end{figure}

In addition to the semantic 3D city model, the framework supports the integration of external urban datasets to enable complementary analyses; other sources may require different storage models than the relational schema used for CityJSON. To demonstrate this, street-network data derived from OpenStreetMap (OSM) for the geographical extent of the CityJSON dataset are stored in a Neo4j graph database, with road junctions modeled as nodes and street segments as edges \citep{sun2026text}. Amenities (e.g., parks, hospitals, etc.) within the same extent are also retrieved and linked to their nearest junctions. The graph schema is shown in Figure \ref{fig:graph-schema}. This graph-based representation supports network-oriented analyses such as shortest-path computation, routing, and proximity queries, and later we demonstrate that the LLM can traverse both backends together to answer queries requiring coordinated retrieval.

\textbf{Extensibility.} Neo4j was chosen mainly for its simple property-graph model and easy integration with street-network OSM data, not because it is the only suitable option. More broadly, this second backend is meant to illustrate the framework's extensibility rather than to be exhaustive, and we treat it as a proof of concept; other sources, such as RDF store, sensor data, vector-based retrieval over regulatory documents (e.g., for checking height-regulation compliance), are left as natural extensions.

\begin{figure*}
    \centering
    \includegraphics[width=0.9\linewidth]{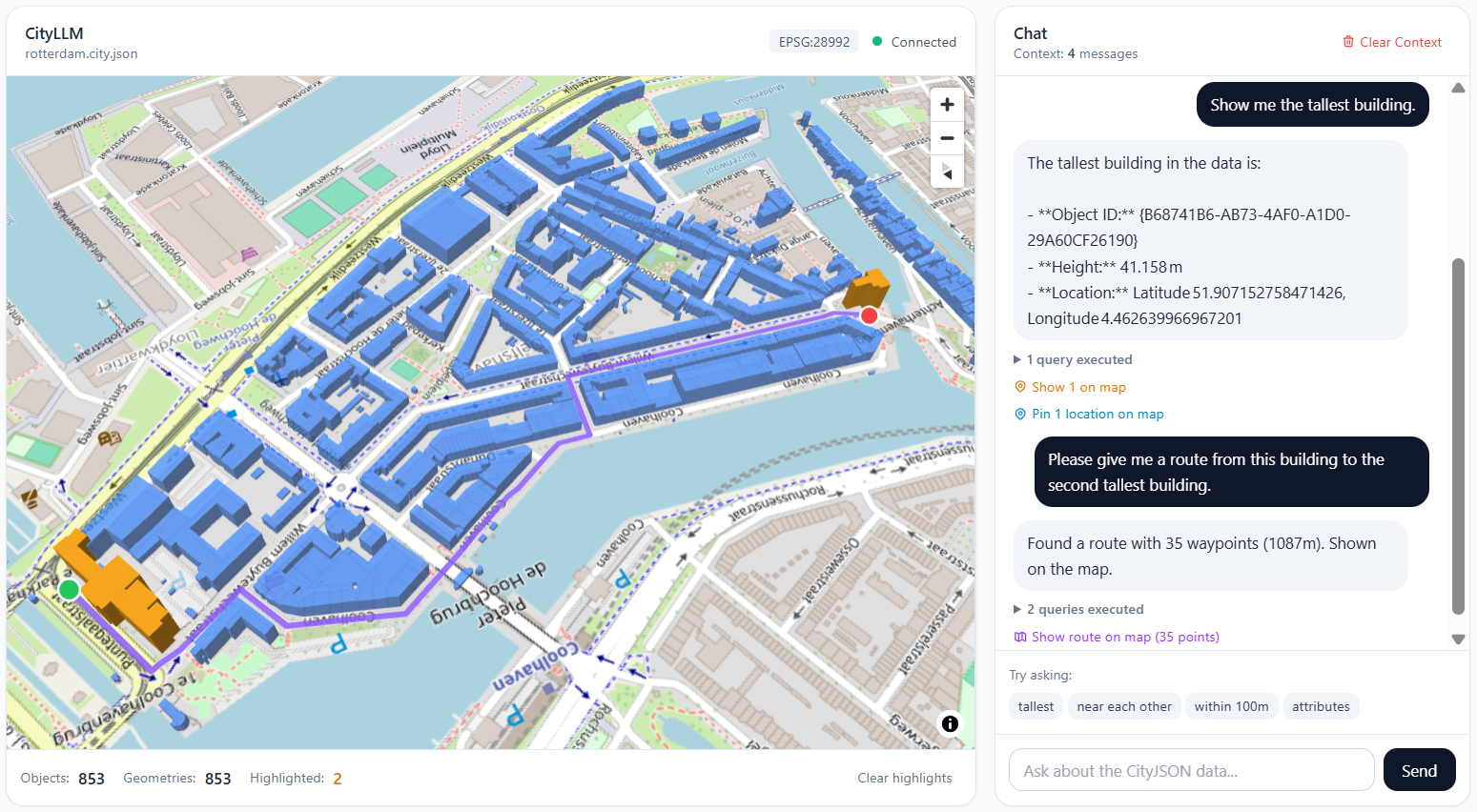}
    \caption{An example conversation with CityLLM.}
    \label{fig:cityllm}
\end{figure*}

\subsection{LLM Agent}
\label{sec:user-query-processing}

User queries in natural-language are processed by an LLM agent that translates them into executable database queries. The agent is provided with the user query together with contextual information about the available data backends, tools, and their schemas via a system prompt. This prompt defines the operational constraints of the agent, including explicit tool selection, structured output format, and query formulation rules. Given the extensive schema, tool descriptions, and rules, the system prompt in CityLLM exceeds 3,000 tokens, with lengths of 3,764 tokens for Gemini and 3,370 tokens for GPT-5.x.

At a high level, the agent selects among three actions: querying PostGIS (CityJSON data), querying Neo4j (OSM street-network graph with amenities), or returning a \textit{cannot\_answer} response for out-of-scope queries. Based on the inferred intent, it generates either SQL (for PostGIS) or Cypher (for Neo4j) queries using the provided schema context. For queries requiring information from multiple data backends, the agent performs \textit{chained execution}, where intermediate results from one query are used to formulate subsequent queries. Table \ref{tab:example-queries} illustrates this for two cross-database examples. In cases where a query execution fails, database error messages are fed back to the LLM, allowing iterative correction and regeneration. This process continues until a valid result is obtained or a predefined \textit{iteration limit} is reached. The final output is then returned as a natural-language response derived from the retrieved data.

\subsection{Interactive Map-based Chat Interface} The framework includes an interactive map interface that allows users to explore query results alongside conversational responses. Depending on the query type, different visualization strategies are applied: (i) building geometries retrieved from PostGIS are highlighted, (ii) point-based results (e.g., amenities) are displayed as points, and (iii) routes computed from the Neo4j graph are rendered as lines. Each response generated by the LLM includes object identifiers and/or coordinates, which are rendered by the frontend. The interface supports incremental updates, allowing users to refine queries and immediately see updated results without resetting the session. This integration of conversational querying with direct visual feedback enables intuitive exploration of semantically rich 3D city data. For transparency, the interface also exposes the SQL and/or Cypher queries generated for each user query. Overall, the framework attempts to bridge natural-language interaction and spatial analysis by integrating LLM-based query generation, database execution, and interactive visualization within a unified workflow.


\begin{table*}
\footnotesize
\centering
\begin{tabular}{lccc}
\toprule
\textbf{Metric} & \textbf{GPT-OSS (120B)} & \textbf{Gemini 3.1} & \textbf{GPT-5.4} \\
\midrule
Answer correctness (expanded in Table \ref{tab:type-results}) & (54/54) 100\% & (51/54) 94.4\% & (46/54) 85.2\% \\
Visualization correctness & (28/28) 100\% & (26/28) 92.9\% & (26/28) 92.9\% \\
Query success & (54/54) 100\% (3 retries) & (54/54) 100\% (2 retries) & (54/54) 100\% (1 retry) \\ 
\bottomrule
\end{tabular}
\caption{Performance comparison across LLMs.}
\label{tab:overall-results}
\end{table*}

\begin{table*}[h]
\footnotesize
\centering
\begin{tabular}{lcccc}
\toprule
\textbf{Query Type} & \textbf{N} & \textbf{GPT-OSS (120B)} & \textbf{Gemini 3.1} & \textbf{GPT-5.4} \\
\midrule
PostGIS & 15 & 15/15 (100\%) & 14/15 (93.3\%) & 11/15 (73.3\%) \\
Neo4j & 12 & 12/12 (100\%) & 12/12 (100.0\%) & 10/12 (83.3\%) \\
Cross-database & 10 & 10/10 (100\%) & 10/10 (100\%) & 10/10 (100\%) \\
Conversational & 17 & 17/17 (100\%) & 15/17 (88.2\%) & 15/17 (88.2\%) \\
\midrule
\textbf{Overall} & \textbf{54} & \textbf{54/54 (100\%)} & \textbf{51/54 (94.4\%)} & \textbf{46/54 (85.2\%)} \\
\bottomrule
\end{tabular}
\caption{Answer correctness by query type for the three evaluated LLMs.}
\label{tab:type-results}
\end{table*}

\section{Evaluation Setup}
\label{sec:evaluation}
The CityLLM framework is implemented in \textit{Python 3.13} using a set of supporting packages. The LLM agent workflow is built with \textit{LangGraph}\footnote{https://github.com/langchain-ai/langgraph}, while the interactive map interface is implemented using \textit{MapLibre GL JS}\footnote{https://github.com/maplibre/maplibre-gl-js}. The street-network and amenity data was retrieved using \textit{OSMnx} \citep{boeing2025modeling}. LLM inference is performed using the following \textit{LangChain integrations}\footnote{https://docs.langchain.com/}: \textit{langchain\_\{groq, openai, google\_genai\}}.

We evaluate CityLLM using the \textit{Rotterdam} dataset available at \url{https://www.cityjson.org/datasets/}. The dataset contains 853 building objects in LoD2. Following the data preparation workflow described in Section~\ref{sec:data-preparation}, the CityJSON dataset was imported into a PostgreSQL/PostGIS database. Additionally, based on the geographical extent (\textit{+ buffer}) of the dataset, a street-network data from OpenStreetMap was stored in a Neo4j graph database. We also extracted amenities (e.g., parks, hospitals) within the same extent and linked each of them to the nearest junction in the Neo4j graph.

\textbf{Evaluation Queries.} Next, we curated a set of $54$ natural-language queries. The queries were constructed to test different functional aspects of CityLLM and were grouped into four types: (i) \textit{PostGIS queries} (N=15): Spatial queries operating on the semantic 3D city model stored in the relational database, (ii) \textit{Neo4j queries} (N=12): Graph-based queries involving street-network relationships, (iii) \textit{Cross-database queries} (N=10): Queries requiring information from both databases, and (iv) \textit{Conversational queries} (5 interactions; N=17): Sequential interactions where subsequent queries depend on previous context. Each query was submitted through the CityLLM interface (as shown in Figure~\ref{fig:cityllm}) and processed as per the workflow illustrated in Figure~\ref{fig:workflow}. 

\textbf{LLMs.} Three LLMs were used: \textit{GPT-OSS (120B)}, \textit{Gemini 3.1 (Flash Lite)}, and \textit{GPT-5.4}. GPT-OSS was selected as a representative open-source LLM suitable for local deployment, while the other two serve as high-performance proprietary baselines. To further examine failure cases, we additionally evaluate model variants (\textit{GPT-5.4-mini} and \textit{Gemini 3.1 Pro}), as discussed later. For all LLMs, the \textit{iteration limit} for query refinement and \textit{chain limit} for cross database querying was set to $3$. 

\textbf{Evaluation Criteria.} The evaluation was done as per following criteria: (i) \textit{Answer correctness}: Whether the final response correctly answered the user’s query, (ii) \textit{Visualization correctness}: For queries requiring spatial output, whether the correct objects or paths were retrieved and accurately visualized, and (iii) \textit{Query success}: Whether a valid executable database query was generated within the allowed retry limit. Among the 54 queries, 28 required visualizations. All three criteria were checked manually by the authors, comparing each response and visualization against the expected result for that query; partial matches were counted as incorrect.

\section{Results and Discussion}
\label{sec:results}

The evaluation results are summarized in Tables \ref{tab:overall-results} and \ref{tab:type-results}. Table \ref{tab:overall-results} presents the performance of the three LLMs across the evaluation criteria, while Table \ref{tab:type-results} provides a breakdown of answer correctness by query type. Across the 54 natural-language queries, GPT-OSS achieved perfect answer correctness. Gemini 3.1 produced incorrect answers in 3 cases, while GPT-5.4 generated 8 incorrect responses. For queries requiring visualization, all LLMs performed strongly: GPT-OSS again achieved perfect accuracy, whereas Gemini 3.1 and GPT-5.4 produced 2 incorrect visualizations. All three models required only a small number of retries across evaluations, indicating that the query refinement mechanism was effective in recovering from initial failures.

\textbf{Discussion.}
The evaluation highlights several key observations. The perfect query success rate across all LLMs indicates that the framework design, particularly iterative query refinement and constrained execution, effectively helps ensure the generation of syntactically valid and executable database queries. As a result, performance differences primarily reflect how accurately each LLM interprets user intent and plans the sequence of operations required to retrieve relevant data.

\begin{figure*}[t]
    \centering
    
    \begin{subfigure}[t]{0.33\textwidth}
        \centering
        \includegraphics[width=\linewidth]{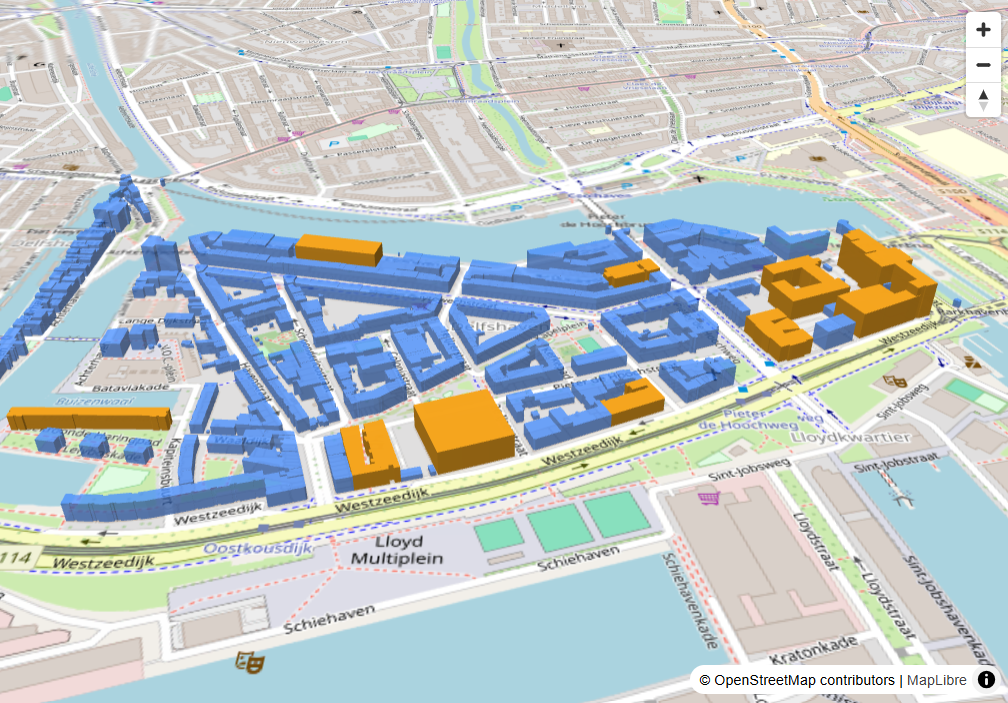}
        \caption{\textbf{GPT-OSS}\\
        \textbf{Query}: \textit{Give me the 10 largest buildings by footprint area.} \\
        \textbf{Response}: \textit{Found 10 results.}}
        \label{fig:oss-visualization-results}
    \end{subfigure}
    \hfill
    \begin{subfigure}[t]{0.33\textwidth}
        \centering
        \includegraphics[width=\linewidth]{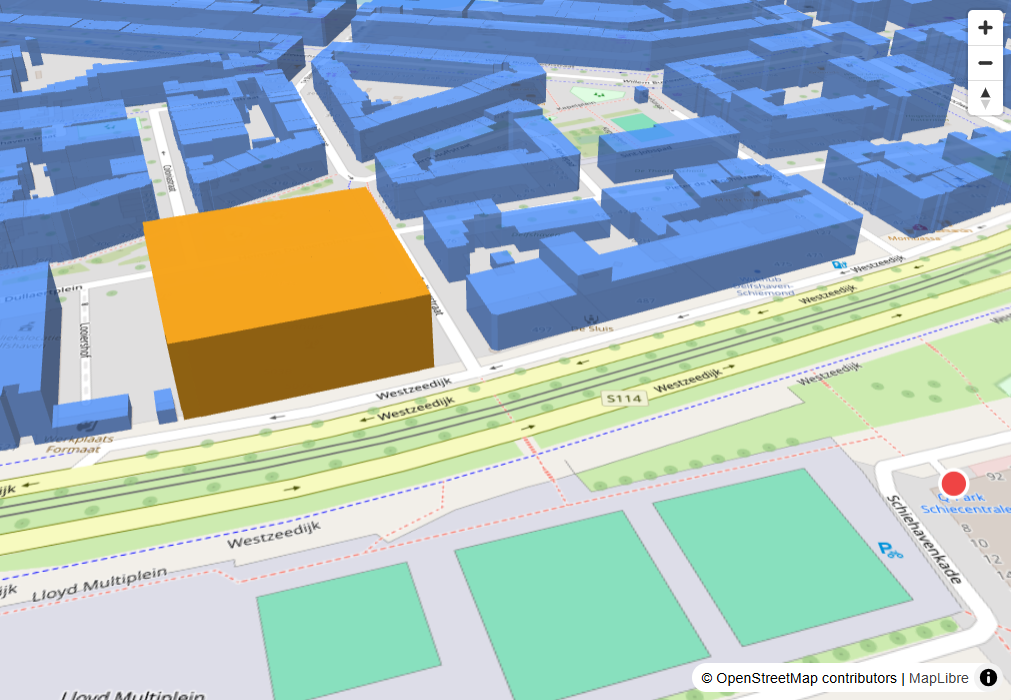}
        \caption{\textbf{Gemini 3.1}\\
        \textbf{Query}: \textit{How far is the nearest parking from the building with the biggest footprint?} \\
        \textbf{Response}: \textit{The nearest parking to the building with the biggest footprint is Q-Park Schiecentrale, located 194.67 meters away.}}
        \label{fig:gemini-visualization-results}
    \end{subfigure}
    \hfill
    \begin{subfigure}[t]{0.33\textwidth}
        \centering
        \includegraphics[width=\linewidth]{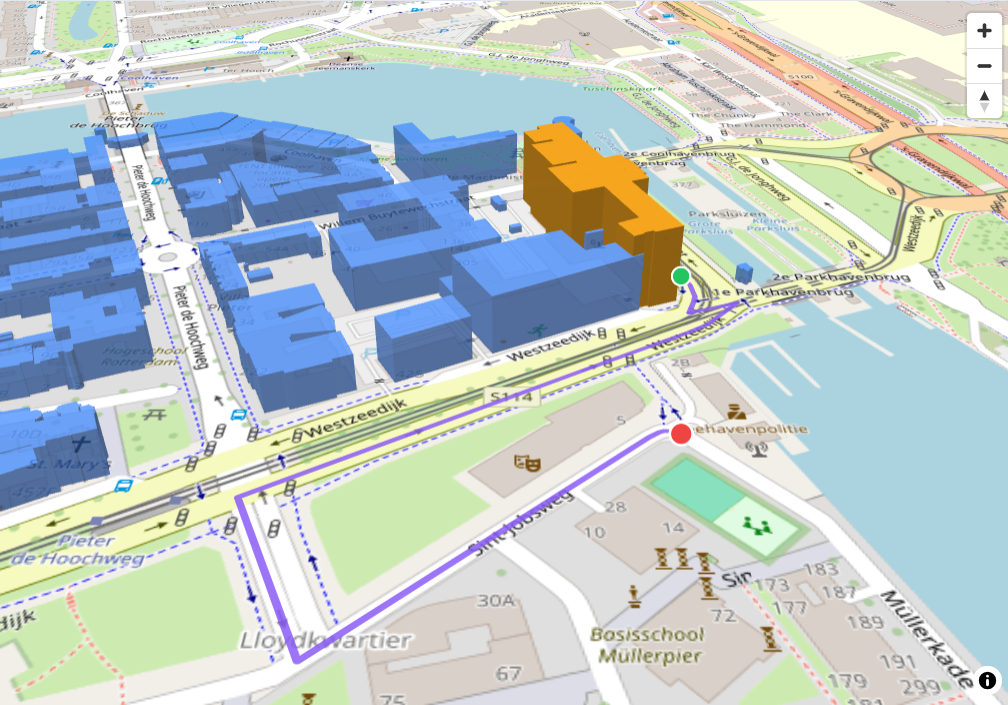}
        \caption{\textbf{GPT-5.4}\\
        \textbf{Query}: \textit{Show me a route from the tallest building to the nearest police station.} \\
        \textbf{Response}: \textit{Found a route with 33 waypoints. Shown on the map.}}
        \label{fig:gpt-visualization-results}
    \end{subfigure}
    
    \caption{Representative visualization outputs produced by the evaluated LLMs.}
    \label{fig:visualization-results}
\end{figure*}

In terms of answer correctness, GPT-OSS achieves perfect performance in this evaluation setting across all query types, suggesting that an open-source LLM can match or even exceed proprietary LLMs under the evaluated setup when guided with structured prompts and detailed schema context. GPT-5.4 shows the lowest overall accuracy (85.2\%), suggesting that even strong reasoning-oriented models may struggle with structured data retrieval tasks. Gemini 3.1 performs more consistently (94.4\%), although some performance loss is observed in PostGIS and conversational queries. Sample success cases and all failure cases are presented in Tables \ref{tab:success-cases} and \ref{tab:failure-cases}, respectively. Additionally, two example cases of cross-database queries are provided in Table \ref{tab:example-queries}. Similarly, selected visualization outputs produced by the evaluated LLMs are presented in Figure \ref{fig:visualization-results}.

\begin{table*}
\footnotesize
\centering
\begin{tabular}{p{2cm} p{4.5cm} p{9.1cm}}
\toprule
\textbf{Query Type} & \textbf{Query} & \textbf{GPT-OSS} \\
\midrule

\multirow{7}{*}{PostGIS}
& \textbf{Q1}: Give me the 10 largest buildings by footprint area.
& Found 10 results. Showing first 5: \texttt{[list]} \\

& \textbf{Q2}: What is the height range of buildings? Like the minimum and maximum. 
& The height range of buildings in the data is 2.2m (minimum) to 41.158m (maximum).\\

& \textbf{Q3}: Show me the location with coordinates of the tallest building. 
& The tallest building in the dataset is located at: \texttt{[lat,lon] coordinates}\\

\midrule

\multirow{9}{*}{Neo4j}
& \textbf{Q1}: How many road junctions are in the network? 
& There are 1731.\\

& \textbf{Q2}: What types of amenities are available? 
& Found 22 results. Showing first 5: \texttt{[list]}\\

& \textbf{Q3}: What's the longest single road segment in the network? 
& The longest single road segment in the network is 202.15 meters long (from junction 12408886628 to junction 4767215171).\\

& \textbf{Q4}: Find the 5 nearest cafes to the first hospital you can find. 
& Hospital: Erasmus Medisch Centrum

5 nearest cafés:
\texttt{[list: Café name | Latitude | Longitude | Distance from hospital]}\\

\midrule

\multirow{5}{*}{Cross-database}
& \textbf{Q1}: How far is the nearest parking from the building with the biggest footprint? 
& The nearest parking (Q‑Park Schiecentrale) is approximately 195 meters from the building with the largest footprint.\\

& \textbf{Q2}: What's the shortest route between the two tallest buildings? 
& Found a route with 35 waypoints (1087m). Shown on the map.\\

\midrule

\multirow{6}{*}{Conversational}
& \textbf{sequential interaction} & \\
& $\rightarrow$ What is the tallest building? 
& The tallest building in the data is the one with ID {B687..}, which is 41.158 meters high.
\\

& $\rightarrow$ And the shortest? 
& The shortest is 2.2m (object ID {C179...}).\\

& $\rightarrow$ How far apart are they? 
& They are approximately 507.36 meters apart.\\

& $\rightarrow$ Give me a route between them. 
& Found a route with 28 waypoints (691m). Shown on the map.\\

\bottomrule
\end{tabular}
\caption{Sample success cases across query types. GPT-OSS responses are shown.}
\label{tab:success-cases}
\end{table*}

\begin{table*}
\footnotesize
\centering
\begin{tabular}{p{2cm} p{4cm} p{4cm} p{2.4cm} p{2.4cm}}
\toprule
\textbf{Query Type} & \textbf{Query} & \textbf{GPT-OSS} & \textbf{Gemini 3.1} & \textbf{GPT-5.4} \\
\midrule

\multirow{14}{*}{PostGIS}
& \textbf{Q1}: What kinds of city objects are there? 
& Building 
&  -
& \textit{[Failed to return distinct types.]} \\

& \textbf{Q2}: How many objects are there for each type? 
& There are 853 objects of type Building.
& -
& \textit{[Failed to aggregate results]} \\

& \textbf{Q3}: List me the attributes available for the buildings. 
& The available building attributes are: bron\_geo, TerrainHeight, bron\_tex, voll\_tex, status.
& -
& \textit{[Failed to return distinct attributes]} \\

& \textbf{Q4}: How many levels of detail are available in the geometries? 
& There are 1.
& \textit{[Returned LoD value, not LoD count.]}
& - \\

& \textbf{Q5}: What are the different roof types and how many buildings have each? 
& No results found for that query. 
& - 
& \textit{[executed cannot\_answer action]}\\

\midrule

\multirow{4}{*}{Neo4j}

& \textbf{Q1}: What types of amenities are available? 
& Found 22 results. Showing first 5: \texttt{[amenities list]}
& -
& \textit{[same as above]}\\

& \textbf{Q2}: How many roads have speed limit available? 
& There are 2387.
& -
& \textit{[same as above]}\\

\midrule

\multirow{16}{*}{Conversational}
& \textbf{\underline{Sequence 1}}: How many hospitals are in the area? $\rightarrow$ Can you list them with their names? $\rightarrow$ \textbf{Which building is closest to the first one?} 
& The building with object ID {2283...} is the closest to the hospital, at a distance of approximately 497m.
& \textit{[executed an SQL with the hospital name in postgres.]} 
& \textit{[same as above]}\\

& & & & \\

& \textbf{\underline{Sequence 2}}: What is the tallest building? $\rightarrow$ And the shortest? $\rightarrow$ How far apart are they? $\rightarrow$ \textbf{Give me a route between them.} 
& Found a route with 28 waypoints (691m). Shown on the map.
& \textit{[Gave a route to a different object.]} 
& - \\

& & & & \\

& \textbf{\underline{Sequence 3}}: Find two nearest cafes to the tallest building. $\rightarrow$ And the shortest? $\rightarrow$ Now show me the route to the first cafe. $\rightarrow$ \textbf{Now show me the route to the second cafe.} 
& Found a route with 54 waypoints (1021m). Shown on the map.
& - 
& \textit{[Could not determine the second cafe coordinates from prior context]} \\

\bottomrule
\end{tabular}
\caption{Failure cases. GPT-OSS responses are shown for reference, while errors in Gemini 3.1 and GPT-5.4 are described explicitly. Where `-', a correct response was provided by the LLM.}
\label{tab:failure-cases}
\end{table*}

The query-type breakdown shows that cross-database queries are handled consistently across all models (100\% in this evaluation). This suggests that once high-level intent is correctly identified, decomposition into sequential sub-queries is reliably managed by the framework pipeline\footnote{Table \ref{tab:example-queries} provides two cases of cross-database chaining.}. In contrast, single-database spatial queries seem to be more sensitive to model-specific reasoning errors. Notably, GPT-5.4 failed on relatively simple aggregation queries such as ``\textit{What kinds of city objects are there?}'', ``\textit{How many objects are there for each type?}'', and ``\textit{List the attributes available for buildings.}'' In these cases, GPT-OSS and Gemini 3.1 correctly applied operations such as \texttt{DISTINCT} and \texttt{GROUP BY}, whereas GPT-5.4 failed to perform the required aggregation or summarization.

Conversational queries introduce different challenges. While GPT-OSS maintains perfect performance, Gemini 3.1 and GPT-5.4 exhibit reduced accuracy (88.2\% for both models), likely due to incomplete or incorrect propagation of context across turns. For example, in the sequence: ``How many hospitals are in the area? $\rightarrow$ Can you list them with their names? $\rightarrow$ \textit{Which building is closest to the first one?}'', GPT-OSS correctly maintained context across all turns. For the final query, it retrieved the coordinates of the first hospital from Neo4j and used them to compute proximity in PostGIS. In contrast, Gemini 3.1 correctly queried Neo4j but propagated the hospital name instead of its coordinates when querying PostGIS, resulting in an empty result set. GPT-5.4, in this case, defaulted to the \texttt{cannot\_answer} action. Similarly, in the sequence ``What is the tallest building? $\rightarrow$ And the shortest? $\rightarrow$ How far apart are they? $\rightarrow$ \textit{Give me a route between them.}'', Gemini 3.1 lost context in the final step and generated a route to an unrelated object. In contrast, both GPT-OSS and GPT-5.4 correctly maintained context and produced the expected route between the identified buildings. These failures in maintaining conversational state ultimately affect not only answer correctness but also visualization accuracy.

\begin{table*}[t]
\footnotesize
\setlength{\tabcolsep}{4pt}
\renewcommand{\arraystretch}{1.05}
\centering
\begin{tabular}{p{0.18\linewidth} p{0.24\linewidth} p{0.53\linewidth}}
\toprule
\textbf{Query} & \textbf{Response} & \textbf{Generated SQL/Cypher} \\
\midrule

Are there any schools close to the smallest building? &

Yes. The nearest school is \textbf{Delfshaven}, about \textbf{167\,m}
from the smallest building (height \textbf{2.2\,m}); other schools are
within 300--400\,m. &

\parbox[t]{\linewidth}{\ttfamily\footnotesize
SELECT object\_id, height\_m,

ST\_Y(ST\_Transform(ST\_SetSRID(centroid,28992),4326)) AS lat, ST\_X(ST\_Transform(ST\_SetSRID(centroid,28992),4326)) AS lon FROM cityllm.cj\_llm\_preferred WHERE obj\_type=`Building' AND height\_m IS NOT NULL ORDER BY height\_m ASC LIMIT 1; \newline

MATCH (a:Amenity) WHERE a.amenity\_type=`school' WITH a, point.distance(point(\{latitude:a.lat, longitude:a.lon\}), point(\{latitude:51.90770095132175, longitude:4.45532072772442\})) AS dist ORDER BY dist LIMIT 5 RETURN a.name AS school\_name, a.lat AS lat, a.lon AS lon, dist AS distance\_m
}\\

\midrule

Find and name 5 parks near the center of the city, using the centroid of all buildings as the center point. &

Centroid of all buildings: \textbf{51.9066269N,4.4543036E.}

5 Nearest parks: \textbf{De Schat van Schoonderloo} (113\,m), \textbf{Kleine Avonturen} (489\,m), \textbf{Zonne Tuin} (548\,m), \textbf{Tuschinskipark} (615\,m), and one unnamed park (806\,m). &

\parbox[t]{\linewidth}{\ttfamily\footnotesize
SELECT `all\_buildings\_centroid' AS object\_id, ST\_Y(ST\_Centroid(ST\_Collect(ST\_Transform(centroid,4326)))) AS lat, ST\_X(ST\_Centroid(ST\_Collect(ST\_Transform(centroid,4326)))) AS lon FROM cityllm.cj\_llm\_preferred WHERE obj\_type=`Building'; \newline

MATCH (a:Amenity) WHERE a.amenity\_type=`park' WITH a, point.distance(point(\{latitude:a.lat, longitude:a.lon\}), point(\{latitude:51.90662691347196, longitude:4.454303621614631\})) AS dist ORDER BY dist LIMIT 5 RETURN a.name AS park\_name, a.lat AS lat, a.lon AS lon, dist AS distance\_m
} \\
\bottomrule
\end{tabular}
\caption{Example natural-language queries with corresponding LLM (GPT-OSS) responses and generated SQL (PostGIS) and Cypher (Neo4j) queries, illustrating cross-database chaining.}
\label{tab:example-queries}
\end{table*}

\textbf{Failure Case Re-evaluation Using Model Variants.}
To further examine the observed errors, we re-evaluated failure cases using alternative variants from the same model families. Specifically, failed responses from Gemini 3.1 (Flash Lite) were re-tested with Gemini 3.1 Pro, and failures from GPT-5.4 were re-evaluated using GPT-5.4-mini. This allows us to assess whether errors are consistent across a model family or specific to a given variant. For Gemini, the Pro variant resolved all previously observed failures. It correctly handled the level-of-detail query by identifying \textit{LoD2} as the only available level, rather than returning a raw value. In the conversational hospital example, this time, it propagated coordinates, thereby performing correct cross-database querying. It also produced the correct route in the routing failure case, although one retry was required. These results suggest that the earlier failures are not inherent to the Gemini family, but are dependent on the specific variant. For GPT-5.4, the mini variant corrected three out of eight failure cases. It successfully handled aggregation-based queries such as identifying object types, counting objects per type, and listing building attributes, by correctly applying \texttt{DISTINCT} and \texttt{GROUP BY}. However, the remaining failures persisted, with the model often defaulting to the \texttt{cannot\_answer} action. This indicates that while some errors are variant-dependent, others appear consistent across the GPT-5.4 family.

\textbf{Implications.}
The results show that CityLLM can effectively provide natural-language interaction capability over semantic 3D city models. The evaluation confirms that strong performance is achieved when LLMs correctly interpret user intent and operate strictly within a well-defined context (e.g., schema definitions, tool selection, and structured output formats) with availability of retrying and chaining abilities. Evaluation also shows that even advanced models can underperform when these constraints are not followed. Overall, the findings position CityLLM as a lightweight,  extensible framework for conversational querying of semantic city data. By mapping CityJSON data into a compact relational schema, exposing executed queries, providing interactive visualizations, and supporting modular integration of additional data sources, the framework enables flexible and interpretable workflows. Importantly, it lowers the barrier to accessing and analyzing complex city data, making it useful for non-expert users and interdisciplinary research as such datasets continue to grow in scale and complexity.

\section{Conclusion}
\label{sec:conclusion}

This paper presents CityLLM, a framework for natural-language querying of semantic 3D city models, which integrates LLMs with spatial and graph data backends. Evaluated on a CityJSON dataset of Rotterdam (853 LoD2 buildings) using 54 queries across spatial, graph, cross-database, and conversational scenarios, the results show strong performance, with GPT-OSS outperforming Gemini 3.1 and GPT-5.4. The findings highlight that LLMs perform best when operating within a well-defined context, supported by retrying and cross-database chaining.

Several directions remain for future work. Introducing additional data sources and backends, such as RDF store, vector database and real-time data, could allow more sophisticated urban analysis. Another direction is reducing the system prompt size (currently $>$3k tokens) through dynamic prompt construction, for example by retrieving only relevant contexts via vector search \citep{lamsal2025query}. Finally, integrating BIM data \citep{lamsal2026ifcllm} could enable unified querying across city models and building-level information.

\section*{Acknowledgements}
This work was supported by the Australian Research Council (ARC) grant: ARC ITRP IH210100048, and was done in collaboration with FrontierSI.

\bibliography{ref}

\end{document}